\documentclass[runningheads]{llncs}

\usepackage{eccv}

\usepackage{eccvabbrv}

\usepackage{graphicx}
\usepackage{booktabs}
\usepackage{multirow}

\usepackage[accsupp]{axessibility}  

\usepackage{hyperref}

\usepackage{orcidlink}

\begin{document}

\title{GSD: View-Guided Gaussian Splatting Diffusion for 3D Reconstruction} 

\titlerunning{View-Guided GS Diffusion for 3D Reconstruction}

\author{Yuxuan Mu\inst{1,2}\thanks{This work was done during an internship at Huawei Noah's Ark Lab}\orcidlink{0000-0001-7132-3155} \and
Xinxin Zuo\inst{2}\orcidlink{0000-0002-7116-9634} \and
Chuan Guo\inst{1,2}\orcidlink{0000-0002-4539-0634}  \and
Yilin Wang\inst{1,2}  \and
Juwei Lu\inst{2}  \and
Xiaofeng Wu\inst{2}  \and
Songcen Xu\inst{2}  \and
Peng Dai\inst{2}\ \and
Youliang Yan\inst{2}  \and
Li Cheng\inst{1}\orcidlink{0000-0003-3261-3533}
}

\authorrunning{Y.~Mu et al.}

\institute{University of Alberta, Edmonton AB T6G 2R3, Canada 
\\
\email{\{ymu3, lcheng5\}@ualberta.ca}
\and
Huawei Noah’s Ark Lab
}

\maketitle

\begin{abstract}
  We present GSD, a diffusion model approach based on Gaussian Splatting (GS) representation for 3D object reconstruction from a single view. Prior works suffer from inconsistent 3D geometry or mediocre rendering quality due to improper representations. We take a step towards resolving these shortcomings by utilizing the recent state-of-the-art 3D explicit representation, Gaussian Splatting, and an unconditional diffusion model. This model learns to generate 3D objects represented by sets of GS ellipsoids. With these strong generative 3D priors, though learning unconditionally, the diffusion model is ready for view-guided reconstruction without further model fine-tuning. This is achieved by propagating fine-grained 2D features through the efficient yet flexible splatting function and the guided denoising sampling process. In addition, a 2D diffusion model is further employed to enhance rendering fidelity, and improve reconstructed GS quality by polishing and re-using the rendered images. The final reconstructed objects explicitly come with high-quality 3D structure and texture, and can be efficiently rendered in arbitrary views. Experiments on the challenging real-world CO3D dataset demonstrate the superiority of our approach. Project page: \href{https://yxmu.foo/GSD/}{https://yxmu.foo/GSD/}
  \keywords{Object Reconstruction, Gaussian Splatting, Guided Diffusion Model}
\end{abstract}

\section{Introduction}
\label{sec:intro}

Given the abundance of image data in the real world, the problem of 3D reconstruction from single-view images has garnered notable attention~\cite{wang_pixel2mesh_2018, xie2020pix2vox++, fan_point_2017, yu2021pixelnerf}. While humans can effortlessly deduce the general object shape and even imagine its texture from unseen views, for computational models, the problem becomes highly non-trivial. As will be described next, three key aspects underpin this problem. First, a proper 3D representation capable of encoding high-fidelity 3D information, while being compatible with various levels of quantization. Second, akin to the human perception system, it is crucial to have a generative model being able to produce an object with diverse appearances of the object's backside and being faithful to the input views. Finally, the ability to efficiently and precisely render a 3D object into an arbitrary view.

\begin{figure}[tb]
\centering
    \includegraphics[width=0.9\textwidth]{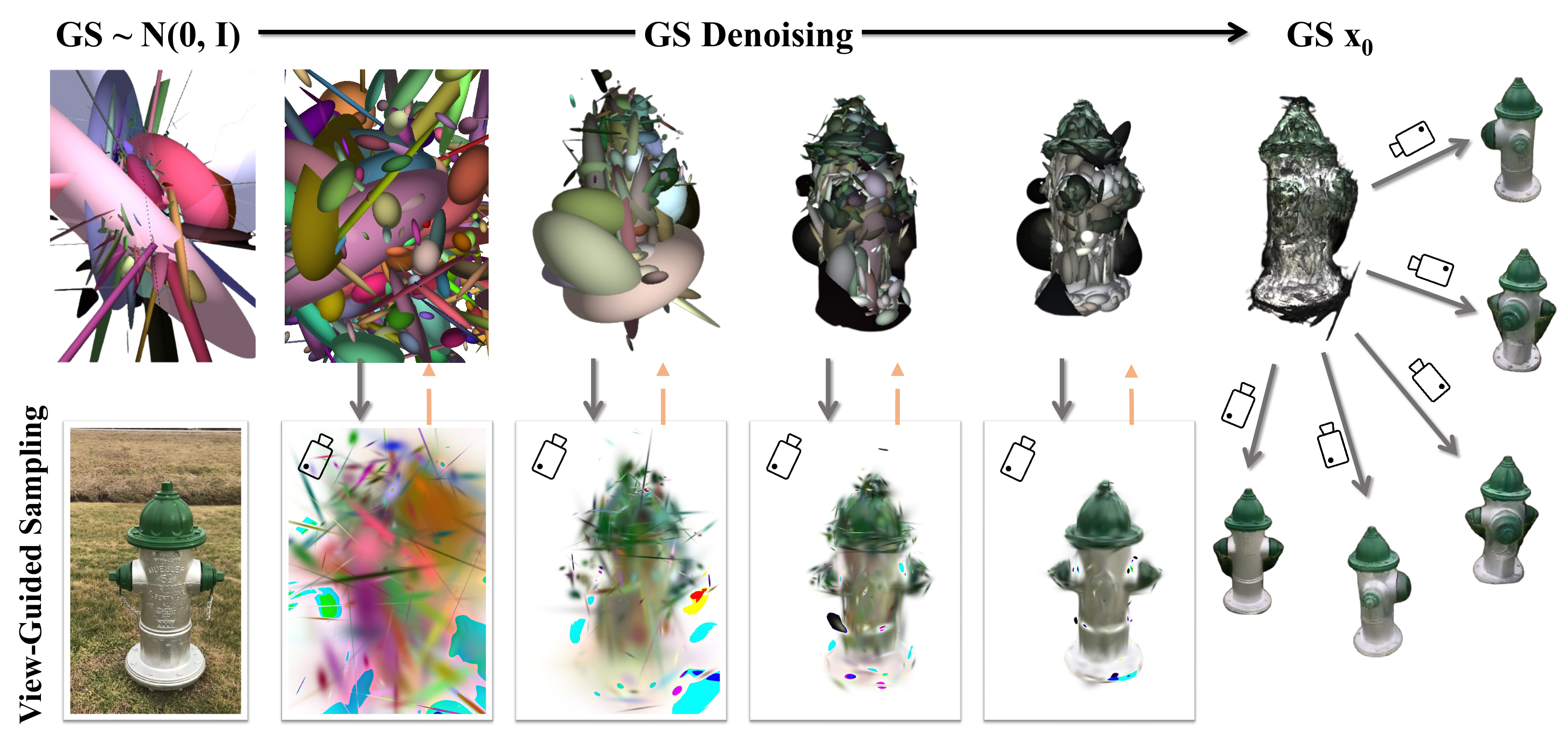}
    \captionof{figure}{A illustration of our \textbf{View-Guided Gaussian Splatting Diffusion} framework for single-view 3D reconstruction. It works by progressively denoising a randomly initialized set of Gaussian Splatting (GS) ellipsoids with continuous guidance from the discrepancies between the input and rendered images. The \textcolor{gray}{gray} arrow represents the splatting-based GS rendering, while the \textcolor{orange}{orange} arrow depicts the backpropagation of guidance gradients. The diffusion model built directly upon the GS representation in our context provides explicit geometry information. The view-guided sampling takes the advantage of splatting function to faithfully yet efficiently obtain fine-grained features from the given view. }
    \label{fig:teaser}
\end{figure}

Existing efforts often struggle to properly address one or multiple aspect(s) of the above three. For instance, the recent 2D novel view synthesis methods~\cite{zhou2023sparsefusion, chan2023generative, liu2023zero, kulhanek2022viewformer, reizenstein2021common, watson2022novel} usually fall short in maintaining 3D consistency. Another line of research~\cite{xie2020pix2vox++, avidan_few-shot_2022, yang_single-view_2021, fan_point_2017, lin2018learning, li_efficient_2018, gao_learning_2020, melas-kyriazi_pc2_2023, wang_pixel2mesh_2018} based on explicit 3D representations such as voxels, point clouds and meshes, are limited to coarse geometry and suboptimal rendering quality, though offering consistent 3D rendering. This may be attributed to the low resolution or sparse nature of 3D features (\eg, points, voxels) and the challenges of their discretization (\eg, mesh) in deep learning models. Implicit 3D representation, on the other hand, formulates 3D space as a query-based implicit function~\cite{chibane_implicit_2020, mildenhall2021nerf, cao2023hexplane, muller2023diffrf, shue20233d}, achieving remarkable quality in single-view 3D reconstruction in well-defined canonical space~\cite{yu2021pixelnerf, chen2023single, jang2021codenerf}. Unfortunately, they typically rely on cumbersome efforts such as marching cubes for 3D geometry extraction and view rendering. 

Motivated by these observations, we introduce a novel framework, GSD, for high-quality single-view 3D reconstruction by building a generative Diffusion Transformer (DiT)~\cite{peebles2023scalable} upon the emerging Gaussian Splatting (GS) representation~\cite{kerbl20233d}. Specifically, GS encodes a scene by a set of GS ellipsoids, with each ellipsoid parameterized by its center position, covariance, regional color, and opacity. Unlike existing 3D representations, GS explicitly encodes 3D geometry and texture in high resolution and density. Furthermore, due to its spatial explicitness, we can easily deploy a point-space DiT even without positional encoding~\cite{nichol2022point-e}. Different from other diffusion models using classifier-free conditioning with image encoders\cite{zhou2023sparsefusion, nichol2022point-e, long2023wonder3d}, our GS DiT enables efficient fine-grained image conditioning through its unique splatting-based rendering and loss-guided sampling~\cite{song2023loss}, which also ensures fidelity to the given-view images.

\begin{figure}[t]
  \centering
  \includegraphics[width=0.8\linewidth]{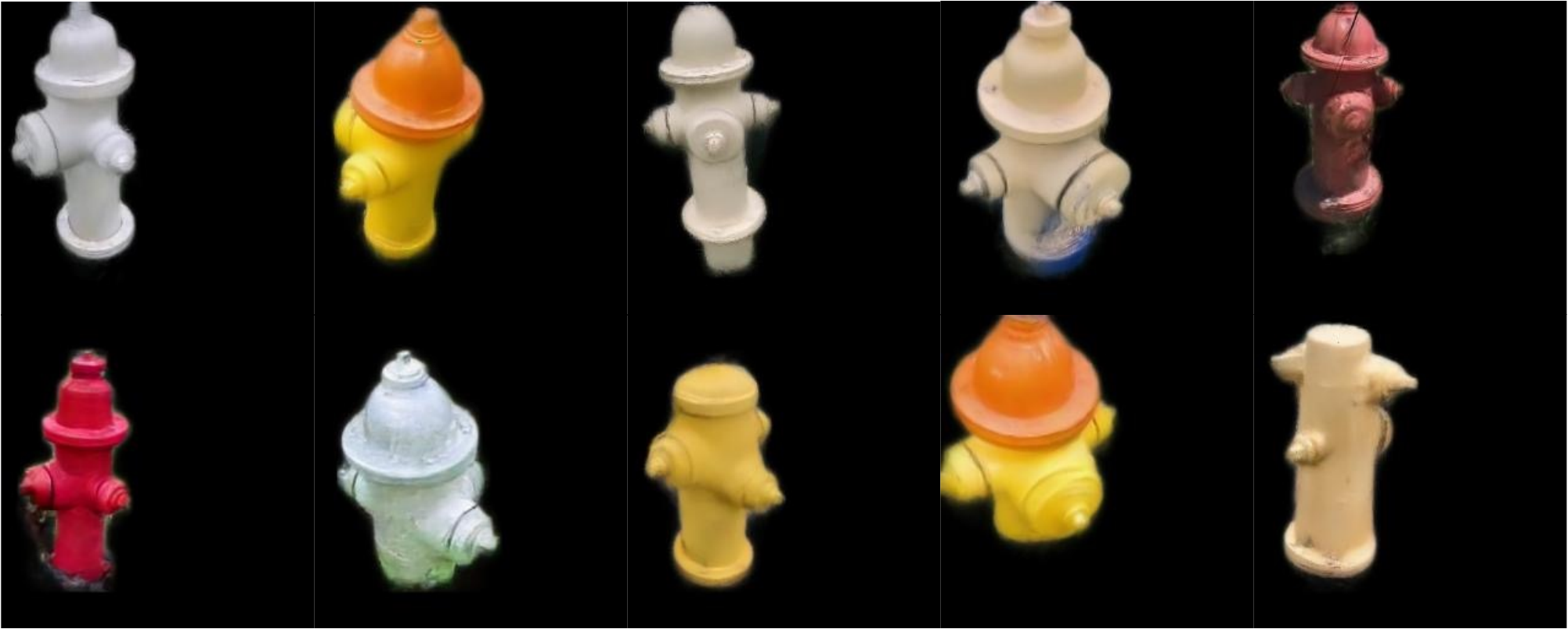}
  \caption{\textbf{Unconditional Generation of GS DiT on the Hydrant dataset.} Ten distinct samples are generated from our unconditional diffusion model which is trained on more than 500 hydrant scenes using the GS representation. Our diffusion model shows an appealing ability on modeling the generative priors of 3D objects.}
  \label{fig:uncondgen}
\end{figure}

Upon the GS representation, a category-specific 3D DiT is trained to capture the space of plausible 3D objects in terms of their diverse geometries and textures. As illustrated in~\cref{fig:uncondgen}, when w/o an input image, our diffusion model learns to generate high-fidelity 3D objects with distinct geometries and textures. When an input image is provided, the same diffusion model is used to reconstruct the specific 3D object, faithful when rendering to the same view. This process is presented in~\cref{fig:teaser}. During test time, the GS object at each denoising step is projected to the given view through the differentiable splatting-based rendering. The gradients of discrepancies between the rendered and the reference images are then backpropagated to the corresponding GS samples to refine the 3D object at the current step, similar to classifier guidance~\cite{song2023loss}. This approach is simple yet effective, and can be easily adapted to multi-view reconstruction. In addition, an auxiliary 2D diffusion model is employed to further improve the quality of rendered images, which reciprocally facilitates better 3D reconstruction.

Our main contributions can be summarized as follows:
\begin{itemize}
    \item Our proposed GSD is, to our knowledge, the first diffusion model that directly models raw GS representation capturing its 3D generative prior for single-view reconstruction. 
    \item The GS DiT intuitively comes with an effective yet flexible view-guided sampling strategy that can extract fine-grained features from the given views using the efficient splatting function.
    Given an input image at test-time, the guided iterative denoising of our GS-based diffusion model allows a progressive refinement of the reconstructed 3D object consistent with the input view. 
    \item Empirical experiments of the real-world CO3D dataset demonstrate the superiority of our approach when compared to the state-of-the-art. Our approach is flexible and can also work with multi-view images.
    
\end{itemize}

\section{Related Work}
\label{sec:relatedwork}

\noindent\textbf{View-Conditioned 3D Reconstruction and Generation.}
Many related works reconstruct the 3D shape by jointly modeling the unconditional 3D priors and the conditional distribution with a generative model~\cite{jun2023shap-e, nichol2022point-e, zeng2022lion}. Some of them working on explicit 3D representations~\cite{zeng2022lion, nichol2022point-e, melas-kyriazi_pc2_2023, xie2020pix2vox++} can recover explicit geometry but fail to synthesize photorealistic views. For the other stream of studies~\cite{jun2023shap-e, chen2023single, muller2023diffrf, shue20233d, cao2023hexplane, chan2023generative}, the use of advanced implicit representations enables photo-quality view synthesis while struggling to extract accurate geometry in unconstrained space. We find the emerging 3D representation Gaussian Splatting~\cite{kerbl20233d} has great potential to be a generally suitable representation for this task, which enjoys both benefits from explicit geometry and realistic view synthesis. While concurrent works using GS representation either re-form a deterministic prediction problem~\cite{szymanowicz2023splatter} or combine with latent representation for additional feature decoration~\cite{zou2023triplanegaussian, xu2024agg, chen2024gssurvey}. Moreover, most of the previous works perceive the given view in the camera space by an image encoder~\cite{nichol2022point-e, jun2023shap-e, tang2021skeletonnet}, which cannot ensure faithful reconstruction due to the compression and also requires canonical coordinates to constrain the modeling space. They indeed should be image-conditioned generation, rather than reconstruction from views. Inspired by the fine-grained projection method from PC$^2$~\cite{melas-kyriazi_pc2_2023}, we take advantage of GS splatting-based rendering~\cite{kerbl20233d} to get access to the image through pixel-level gradients that reliably keep the view information and are flexible to accommodate arbitrary views using relative camera parameters in world space. 
A similar gradient conditioning approach is also used in SSD NeRF~\cite{chen2023single}, while restricted by the robustness of its dataset-specific neural rendering. Combining GS universal rendering with view-guided sampling conditioning on the GS diffusion model could fully explore the potential of these methods. 

\noindent\textbf{Novel View Synthesis.}
The current novel view synthesis (NVS) task becomes progressively close to the 3D reconstruction task~\cite{zhou2023sparsefusion, rombach2021geometry, kulhanek2022viewformer, reizenstein2021common}. One of the most significant pinpoints is the 3D inconsistency issue from being short of 3D geometry priors. To address this problem, some works try to involve multi-view geometry~\cite{zhou2023sparsefusion, yu2021pixelnerf}, depth~\cite{cao2022fwd, long2023wonder3d}, and clues from large multi-view 2D dataset~\cite{long2023wonder3d, liu2023zero, rombach2021geometry, watson2022novel}. However, since it primarily focuses on imaging geometry awareness rather than modeling the prior distribution of 3D shapes. We argue this 2D objective deviates from 3D reconstruction, potentially resulting in unsatisfactory 3D geometry.


\noindent\textbf{SDS-based 3D Asset Creation.}
3D asset creation emerges thanks to the boom of big models. Most of them distill the 3D representation from a pre-trained image generation model by Score Distillation Sampling (SDS)~\cite{poole2022dreamfusion, tang2023dreamgaussian, chen2023gsgen}. One of the primary weaknesses lies in 3D geometry, despite involving multi-view geometry to regress the 3D representation. These approaches heavily rely on the consistent performance of large pre-trained image models, where views are treated as independent during pre-training. This could lead to the Janus problem. Another issue arises when applying these methods to our object reconstruction scenario, as the 3D space should ideally be well-constrained. However, achieving alignment in a real-world setting is challenging. Consequently, the 3D asset creation approaches may not easily adapt to practical applications in real-world object reconstruction.


\section{Background: Gaussian Splatting}
\label{sec:bg}

Gaussian Splatting~\cite{kerbl20233d} presents an emerging method in the field of novel view synthesis and 3D reconstruction from multi-view images. In contrast to NeRF style implicit representations~\cite{mildenhall2021nerf}, GS takes a different approach that characterizes the scene using a set of anisotropic GS ellipsoids defined by their center positions \(\mu \in \mathbb{R}^3\), covariance \(\Sigma \in \mathbb{R}^6\), color \(c \in \mathbb{R}^3\), and opacity \(\alpha \in \mathbb{R}^1\). During rendering, the GS is projected onto the imaging plane and then allocated to individual tiles \cite{zwicker2002ewa}. The color of \(\textbf{p}\) on the image is given by typical point-based blending~\cite{yifan2019splatting} as follows:
\begin{gather}
    \label{formula:splatting}
    C(\textbf{p}) = \sum_{i \in \mathcal{N}} c_{i}\sigma_{i} \prod_{j=1}^{i-1} (1-\sigma_{i}),\\
    \text{where}\ \sigma_{i}=\alpha_{i}e^{-\frac{1}{2}(\textbf{p}-\mu_{i})^{T} \Sigma_{i}^{-1} (\textbf{p}-\mu_{i})} \nonumber.
\end{gather}
Compared with NeRF's importance sampling process, these GS points have the potential to cluster towards critical regions, thereby improving overall efficiency for expression and rendering.

\section{Our Approach}
\label{sec:method}

\begin{figure}[tb]
  \centering
  \includegraphics[width=\linewidth]{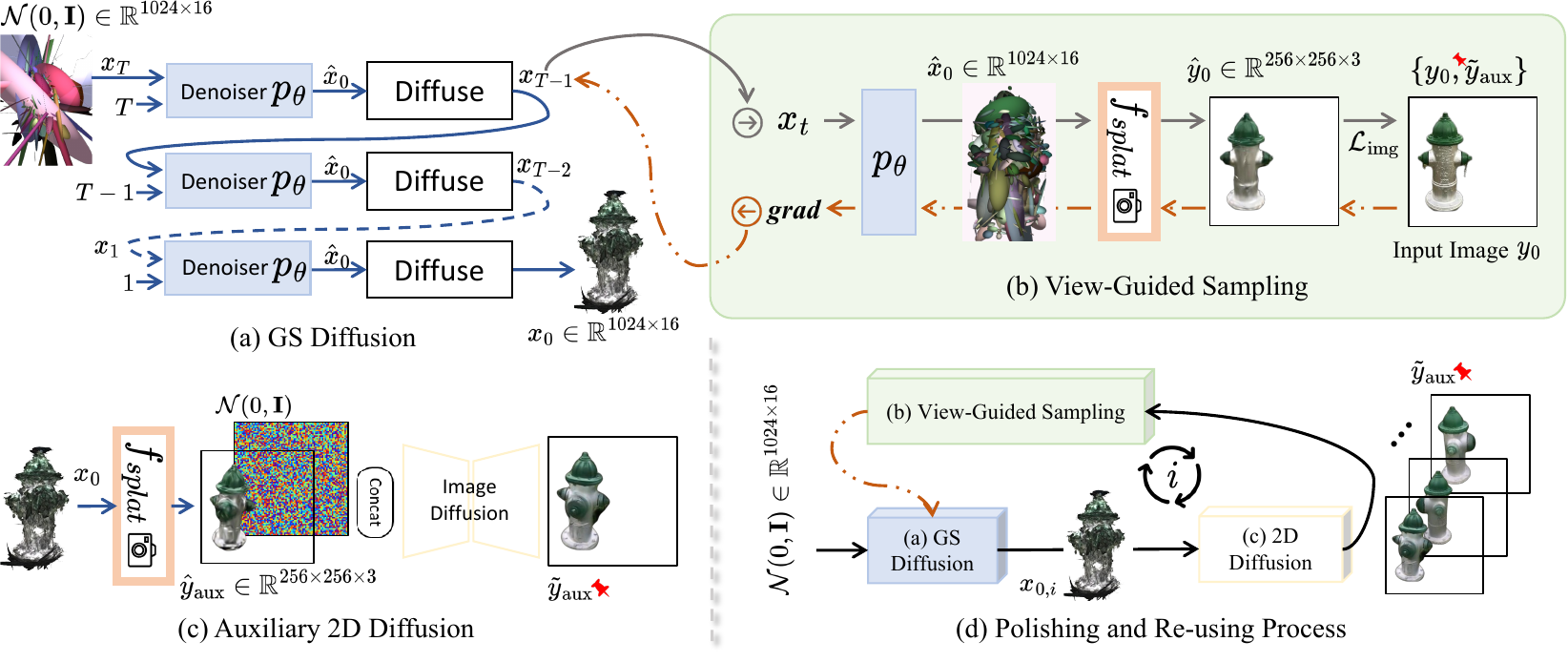}
  \caption{\textbf{Approach Overview.} (a) An unconditional diffusion model is trained on objects represented by N GS ellipsoids (N=1024). After training, the GS ellipsoids of an object can be generated through $T$ denoising steps (\cref{subsec:GSd}). (b) At inference time, we apply view-space loss guidance at each denoising step. The \textcolor{gray}{gray} arrow represents the splatting-based GS rendering, while the \textcolor{orange}{orange} arrow depicts the backpropagation of guidance gradients. The GS object rendering through the splatting function $f_{\text{splat}}$ from input-view is compared with the given image using $\mathcal{L}_{img}$, and the gradients backpropagate to the diffusion model for adjusting the sampling process (\cref{subsec:guidedsample}). (c) A 2D diffusion model is employed to enhance the fidelity of rendered views from reconstructed GS $x_0$. (d) The refined synthetic view images are then re-used to improve GS reconstruction quality in an alternating iterative enhancement manner (\cref{subsec:3d2d}). We obtain the final reconstructed GS object $x_0$ from the last run of GS diffusion.}
  \label{fig:pipeline}
  
\end{figure}

\Cref{subsec:GSd} explains how we formulate the GS diffusion model. In~\Cref{subsec:guidedsample}, we elaborate on how the GS splatting function is utilized to guide the denoising sampling process of GS DiT for view-based 3D reconstruction. Finally, \Cref{subsec:3d2d} details the polishing and re-using process of the auxiliary 2D diffusion model jointly with GS DiT. See~\Cref{fig:pipeline} for an overview of our pipeline at inference.

\subsection{Modeling GS Generative Prior}
\label{subsec:GSd}

Building upon the recent advancements in denoising diffusion probabilistic models (DDPM)~\cite{ho2020denoising}, we formulate our GS dataset distribution modeling using a diffusion-based generative model.


\textbf{Representing 3D Objects with GS.}
To prepare the dataset for training gs diffusion model, we convert our dense-view image dataset into a dataset of GS using the 3D GS scene reconstruction method~\cite{kerbl20233d}. However, this dense-view regression-based approach regularizes the optimization process only through its densify-and-prune function on points. In contrast, for applicable feed-forward network modeling, we aim to further regularize the feature distribution of GS and obtain a constant quantity of GS ellipsoids per-scene. Therefore, we initially restrict the number of GS ellipsoids by densifying the GS ellipsoids only with the Top-K gradients values, where K is the difference between the pre-set maximum ellipsoid quantity and the current ellipsoid quantity. We observe that this constrained densification allows for effective reconstruction of object data with only a 2\% PSNR deduction compared to the full model which has two orders of magnitude more GS ellipsoids. 




\textbf{Training a Diffusion Model on GS.}
In general, a diffusion model takes Gaussian noise as input and progressively denoises it in $T$ steps. It learns strong data priors from the denoising-diffusion process~\cite{zeng2022lion}. In our framework, the diffusion model operates on GS ellipsoids $\mathbf{x} \in \mathbb{R}^{16}$, with features including position $\mathbb{R}^3$, scale of covariance $\mathbb{R}^3$, rotation of covariance $\mathbb{R}^6$ \cite{zhou2019continuity}, opacity $\mathbb{R}^1$ and color $\mathbb{R}^3$. In our GS DDPM, $\mathbf{x}_{T} \sim \mathcal{N}(\mathbf{x}_{T}; 0, \mathbf{I})$ is the purely noisy GS ellipsoids, and $\mathbf{x}_{0} \sim q(\mathbf{x}_{0})$ is a data point sampled from the data distribution, which is practically our GS dataset. 

During training, the diffuse process is described as
\begin{equation}
    q(\mathbf{x}_{t} | \mathbf{x}_{t-1}) = \mathcal{N}(\mathbf{x}_{t};\sqrt{1-\beta_t}\mathbf{x}_{t-1}, \beta_t \mathbf{I}),
\end{equation}
which formulates a Markov process with a variance schedule $\{\beta_t\}^{T}_{t=0}$ that gradually adds Gaussian noise to the GS. The training objective is to learn the reverse denoising process with a neural approximator $p_{\theta}(\mathbf{x}_{0};\mathbf{x}_{t}, t)$ following~\cite{ramesh2022hierarchical}, given by
\begin{equation}
    \label{formula:lddpm}
    \mathcal{L}_{\text{DDPM}} = \mathbb{E}_{ \mathbf{x}_{0}\sim q(\mathbf{x}_{0}), t \sim [1, T]} \left[ \left\|\mathbf{x}_{0}-p_{\theta}(\mathbf{x}_{0};\mathbf{x}_{t}, t) \right\| ^{2}\right].
\end{equation}

\textbf{Backbone Choice for $p_{\theta}$.}
When our neural approximator $p_{\theta}$ operates on GS, it treats GS as point clouds with rich features. To keep it simple, inspired by~\cite{nichol2022point-e, peebles2023scalable}, we employ a vanilla transformer for unconditional GS modeling. 
The transformer acts as a densely-connect Graph Neural Network with implicit edges realized by multi-head attention.
It is possibly more effective at handling this informative unstructured representation compared with other point cloud learning architectures, such as PVCNN~\cite{liu2019point}, which is discussed in \cref{subsec:abla}. 
We choose not to include positional encoding as our explicit GS representation already incorporates positional information. This design allows our architecture to be versatile, accommodating an ideally arbitrary number of GS points. For training efficiency, we keep the number of points fixed at 1024 for category-specific experiments on CO3D~\cite{reizenstein2021common}. We also explore the scaling-up performance of our GS diffusion transformer by training a single model on relative general objects set, OmniObject3D~\cite{wu2023omniobject3d}, with results shown in \cref{fig:omniresults}.

\subsection{View-Guided Sampling}
\label{subsec:guidedsample}

Taking inspiration from the projection conditioning in~\cite{melas-kyriazi_pc2_2023}, the fine-grained conditioning is a more faithful approach compared to global conditioning using an image encoder. Empirical experiments reveal that the naive point projection method fails to effectively convey features from photographs to the informative GS, which is discussed in~\cref{subsec:abla}. 

Considering that GS features on the denoising process can be seamlessly projected to the image space through the splatting function, it is evident that its reverse operation has the potential to backpropagate fine-grained view information to the GS space through gradients. This idea enlightens us on the approach of loss-guided sampling for the diffusion model~\cite{ho2022video, song2023loss}. 

In conditional generation, we may want to draw samples \(x_0\) from the prior distribution subject to certain conditions \(y\). For diffusion models, the conditional score at time \(t\) can be obtained via Bayes’ rule:
\begin{equation}
    \nabla_{x_t} \log p_t(x_t|y) = \nabla_{x_t} \log p_t(x_t) + \nabla_{x_t} \log p_t(y|x_t),
\end{equation}
where the first term is the unconditional score function $\nabla_{x_t} \log p_\theta(x_t)$ learned via the denoising-diffusion objective \cref{formula:lddpm}. For the second term, the naive solution is to train a classifier $p_\phi(y|x_t)$ on paired data \((y, x_t)\) that operates as this posterior distribution, \ie classifier guidance \cite{dhariwal2021diffusion}. However, a labeled dataset for noisy samples is not always available nor flexible. Diffusion Posterior Sampling (DPS)~\cite{chung2022diffusion} instead approximates $p_t(y|x_t)$ by $p_t(y|\hat{x}_0)$, when assuming $p(y|x_0)$ is given, where $\hat{x}_0$ is essentially a point estimation from the denoiser $p_\theta$ in our case. Reconstruction guidance~\cite{ho2022video} simplifies this approximation by assuming $p(y|x_0)$ is Gaussian. So, the $p_t(y|x_t)$ becomes $\mathcal{N}\left[p_\theta(x_t), \left(\bar\beta/ (1-\bar\beta)\right)\mathbf{I}\right]$, \cref{formula:guidancegaussian1}. Loss Guidance~\cite{song2023loss} promotes this method to a more common setting where we have a differentiable loss function $\ell_y$ to replace the MSE estimation, by the following:

\begin{align}
    \text{DPS}(x_t, y) :&= \nabla_{x_t} \log p_t(y|\hat{x}_0) , \label{formula:guidancegaussian0}\\
    &= \nabla_{x_t} -\frac{1-\bar{\beta_t}}{2\bar{\beta_t}} ||x_0- \hat{x}_0 ||^2 ,\label{formula:guidancegaussian1}\\
    &= \nabla_{x_t} -\frac{1-\bar{\beta_t}}{2\bar{\beta_t}} \ell_y(\hat{x}_0). \label{formula:guidance}
\end{align}

Since we only access the noiseless input 2D image $y_{0}$, we utilize the approximator $p_{\theta}(\mathbf{x}_{0};\mathbf{x}_{t}, t)$ and splatting function $f_{splat}$~\cref{formula:splatting}, to compute $\hat{x}_0$ and $\hat{y}_0$, forming a differentiable loss function in~\cref{formula:guidance}, It then approximates the gradients w.r.t. $\mathbf{x}_{t}$, defined as:
\begin{align}
    \textit{grad} &\leftarrow \nabla_{\mathbf{x}_{t}} -\frac{1-\bar{\beta_t}}{2\bar{\beta_t}} 
    (\mathcal{L}_\text{img} \circ f_{splat}) \left(x_{0}, \hat{x}_0\right),\\
    &\leftarrow \nabla_{\mathbf{x}_{t}} -\frac{1-\bar{\beta_t}}{2\bar{\beta_t}} 
    \mathcal{L}_\text{img}\left[y_{0}, f_{splat}\left(p_{\theta}(\mathbf{x}_{0};\mathbf{x}_{t}, t)\right)\right].
\end{align}
where the view-point camera $\textbf{P}$ is omitted in the splatting function $f_{splat}(y; x, \textbf{P})$ for simplicity. The guidance gradients then bias the unconditional score prediction by
\begin{equation}
\label{formula:guide}
    \tilde{\mathbf{x}}_{t} \leftarrow \hat{\mathbf{x}}_{t} + \lambda_{\text{gd}}\frac{\bar\beta}{\sqrt{1-\bar\beta}}\textit{grad},
\end{equation}
where $\lambda_{\text{gd}}$ is empirically a large weighting factor~\cite{ho2022video}. We further perform predictor-corrector sampling~\cite{song2020score} during the denoising approximation by inserting extra Langevin correction steps between DDIM steps~\cite{song2020ddim}.

\subsection{Polishing and Re-using}
\label{subsec:3d2d}
Experiments from concurrent research working on GS~\cite{kerbl20233d, tang2023dreamgaussian, chen2023gsgen, szymanowicz2023splatter} indicate that when the available views are deficient, needle-shaped artifacts may occur in NVS. We also observe that mild perturbation in the GS space would lead to strong annoying needle-shaped artifacts in 2D views, while the 3D geometry is relatively satisfactory. So, we suggest the presence of a domain gap between the GS rendering views and real images. The 3D modeling process may encounter challenges in assigning adequate importance to features crucial for 2D appearance, such as covariance.

Building on the aforementioned hypothesis, we propose to construct an auxiliary 2D diffusion model that takes imperfect GS rendering images as a condition and generates clean, photorealistic images. To further enhance the view rendering quality of GS diffusion reconstruction, we polish and re-use the rendered and refined images by iteratively performing GS diffusion and 2D diffusion, depicted in~\cref{fig:pipeline} (d). 

\section{Experiments}
\label{sec:experiments}

\subsection{Experimental Setup}
\noindent\textbf{Dataset.} We conduct experiments on CO3Dv2~\cite{reizenstein2021common}, an unconstrained multi-view dataset of real-word objects. The dataset is extraordinarily challenging~\cite{zhou2023sparsefusion, melas-kyriazi_pc2_2023, watson2022novel} since it is captured in-the-wild without any coordinate calibration, which is closer to the daily application conditions. We use the dataset-split annotation from \textit{fewview-dev} for training and evaluation. We show results for core-10 categories: hydrant, bench, donut, teddy bear, apple, vase, plant, suitcase, ball and cake. For scaling-up performance, we additionally illustrate qualitative results on general objects with a model trained on OmniObject3D~\cite{wu2023omniobject3d}, which comprises 6,000 scanned objects in 190 daily categories.


\noindent\textbf{Baselines.} We compare our approach against the current state-of-the-art methods: \textit{NerFormer}~\cite{reizenstein2021common}, \textit{ViewFormer}~\cite{kulhanek2022viewformer} and \textit{SparseFusion}~\cite{zhou2023sparsefusion}. We re-train \textit{NerFormer} on each category using its official implementation. For \textit{ViewFormer} which trained across categories, we use their checkpoint for all categories of CO3Dv2. We compare against \textit{SparseFusion} only for reconstruction from two views, since their design doesn't support single view setting. We use the category-specific model provided by the authors. For comparison in 3D geometry, we use the official released checkpoint on hydrant from PC$^2$~\cite{melas-kyriazi_pc2_2023}.


\noindent\textbf{Metrics.} Following prior works, we report standard image metrics: PSNR, SSIM, and LPIPS, that cover different aspects of image quality for evaluation in 2D views. For 3D geometry, we measure F-score@0.01 \cite{tatarchenko2019what3d} and Chamfer Distance. F-score@0.01 evaluates the precision and recall with a threshold 0.01. The reconstructed point with its nearest distance to the ground truth point cloud under the threshold would be considered as a correct prediction.


\noindent\textbf{Implementation Details.} For the diffusion model, we schedule 1000 steps for GS and 500 steps for 2D, both set to predict the clean sample at each step. We build category-specific transformer encoders for GS denoiser $p_{\theta}$ each with 19.6M parameters. The models are trained with GS points number fixed at 1024, for 200k iterations. We mask out the points with outlier scaling features to stabilize the training. We also find L1 loss performs better than MSE in \cref{formula:lddpm} for GS. For the 2D diffusion model, we build a naive UNet denoiser following a common setting with FP16, input size of 256x256, trained for 200k iterations. If not specified otherwise, we use AdamW with default parameters and a learning rate of 0.0001 for optimization.

We take 100 and 25 DDIM steps in the generative sampling process for GS diffusion and 2D diffusion respectively. We use three polishing and re-using  iterations for single-view reconstruction and two for 2-view reconstruction. We adopt multi-view GS refinement initiated from our reconstructed GS, with 2D refined views. We jointly perform this regression and 2D view refinement in an iterative improvement manner for two iterations in the single-view setting. Reconstructing a single instance takes around 3 minutes on an A100 GPU, depending on number of iterations. The object represented by GS is obtained as the final output.

\subsection{Reconstruction on Real-World Images}
We present category-specific reconstruction results from a single viewpoint for objects such as hydrant, donut, teddy bear, and bench. These categories vary in structural complexity, scale, and the captured environment. For each instance, we adopt a similar experiment setting as \cite{zhou2023sparsefusion} that loads 32 linearly spaced views, from which we randomly select 1 input view and assess the performance on the remaining 31 unseen views. 

\begin{table}[tb]
\centering

\resizebox{\linewidth}{!}{ 
\begin{tabular}{lccccccccccccccccccccccc}
\toprule
\multirow{2}{*}{Methods}             & \multicolumn{2}{c}{Hydrant} & \multicolumn{2}{c}{Bench} & \multicolumn{2}{c}{Donut} & \multicolumn{2}{c}{Teddy bear} & \multicolumn{2}{c}{Apple} & \multicolumn{2}{c}{Vase} & \multicolumn{2}{c}{Plant } & \multicolumn{2}{c}{Suitcase} & \multicolumn{2}{c}{Ball} & \multicolumn{2}{c}{Cake} &\multicolumn{3}{c}{All}\\ \cmidrule(l){2-3} \cmidrule(l){4-5} \cmidrule(l){6-7} \cmidrule(l){8-9} \cmidrule(l){10-11} \cmidrule(l){12-13} \cmidrule(l){14-15} \cmidrule(l){16-17} \cmidrule(l){18-19} \cmidrule(l){20-21} \cmidrule(l){22-24}
                    & \small PSNR   & \small LPIPS   & \small PSNR   & \small LPIPS  & \small PSNR  & \small LPIPS  & \small PSNR   & \small LPIPS & \small PSNR   & \small LPIPS & \small PSNR   & \small LPIPS & \small PSNR   & \small LPIPS & \small PSNR   & \small LPIPS & \small PSNR   & \small LPIPS & \small PSNR   & \small LPIPS  &\small PSNR   & \small LPIPS  &\small SSIM \\ \midrule              
NerFormer \cite{reizenstein2021common}   & 16.9   & 0.33   & 15.4   & 0.46  & 18.0   & 0.37   &  13.1    & 0.46  & 18.2 & 0.36 & 16.9 & 0.35 & 16.4 & 0.46 & 19.5 & 0.41 & 16.1 & 0.37 & 16.2 & 0.47 &  16.67 &  0.404 &  0.562\\
ViewFormer \cite{kulhanek2022viewformer} & 16.6   & 0.24   & 15.7   & 0.34  & 17.1   & 0.37   &  12.9    & 0.34  & \underline{19.1} & \underline{0.30} & 17.9 & 0.26 & 15.9 & 0.37 & 20.2 & \underline{0.30} & 16.5 & 0.33 & 16.8 & 0.36 &  16.87 &  0.321 &  0.625\\
Ours  w/o iter                           & \underline{18.7}   & \underline{0.25}   & \underline{15.7}   & \underline{0.32}  & \underline{18.5}   & \underline{0.33}   &  \underline{16.3}    & \underline{0.35}  & 18.6 & 0.30 & \underline{19.4} & \underline{0.23} & \underline{17.4} & \underline{0.33} & \textbf{21.0} & \textbf{0.29} & \underline{17.1} & \underline{0.31} & \underline{18.0} & \textbf{0.32} & \underline{18.07} &  \underline{0.303} &  \underline{0.679}\\
Ours                                     & \textbf{19.7}  & \textbf{0.20}   &  \textbf{16.2} & \textbf{0.31} & \textbf{18.9}  & \textbf{0.32}  &  \textbf{16.8} &  \textbf{0.31} & \textbf{19.2} & \textbf{0.27} & \textbf{20.1} & \textbf{0.22} & \textbf{17.4} & \textbf{0.30} & \underline{20.4} & 0.31 & \textbf{17.7} & \textbf{0.30} & \textbf{18.5} & \underline{0.34} &  \textbf{18.49} &  \textbf{0.288} &  \textbf{0.696} \\
\bottomrule
\end{tabular}
}
\caption{\textbf{Quantitative comparison in view quality of from single-view reconstruction.} We report PSNR $\uparrow$ and LPIPS $\downarrow$ averaged across the testing set. \textbf{Bold} face indicates the best result, while \underline{underscore} refers to the second best. ``iter'' refers to our iterative polishing and re-using strategy.
}
\label{tab:1view}
\end{table}

\begin{figure}[tb]
  \centering
  \includegraphics[width=\linewidth]{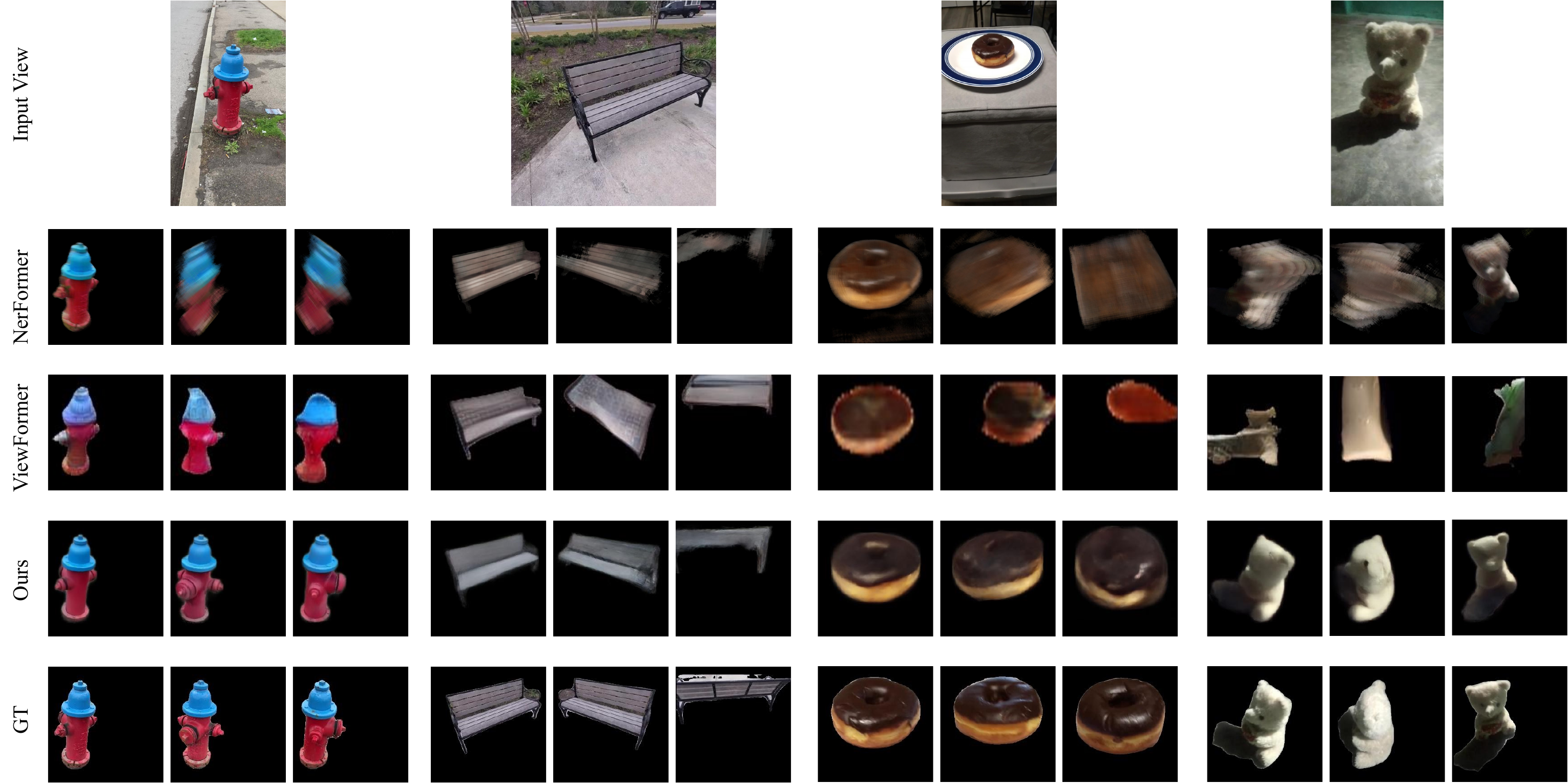}
  \caption{\textbf{View synthesis qualitative results from single-view reconstruction.} We show novel view synthesis results given the object reconstructed from single-view input on hydrant, bench, donut, and teddy bear. Our method takes the raw input view with an object mask with various resolutions as input. Notably, our novel views are rendered from the GS in real-time once we obtain this reconstructed 3D representation.}
  \label{fig:results1view2d}
\end{figure}

\begin{figure}[t]
  \centering
  \includegraphics[width=\linewidth]{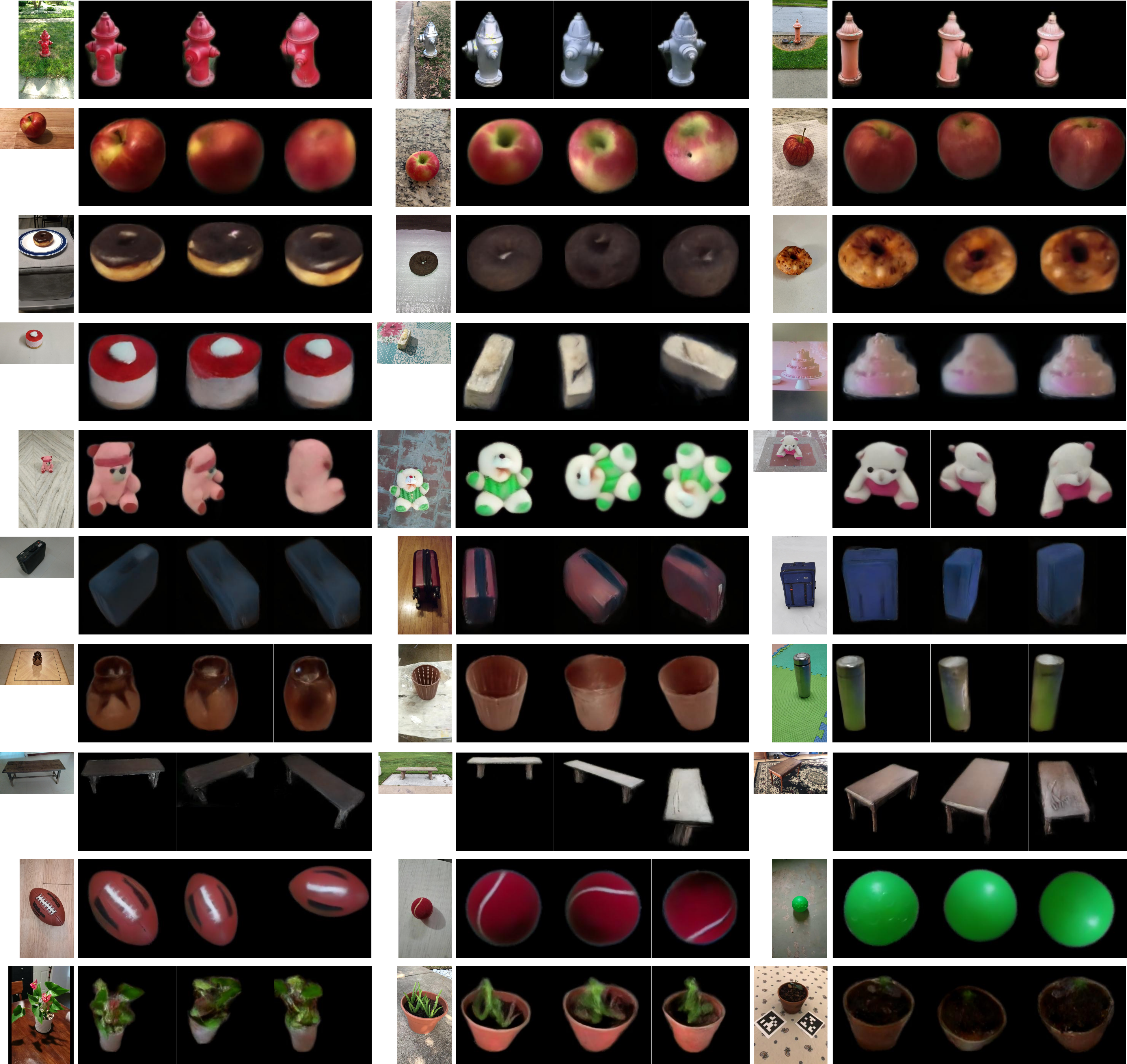}
  \caption{\textbf{Additional view synthesis qualitative results from single-view reconstruction}. We show three novel views rendered from the object reconstructed from single-view input shown on the left.}
  \label{fig:results1view_supp}
\end{figure}

\begin{table}[tb]
\centering
\resizebox{\linewidth}{!}{ 
\begin{tabular}{lcccccccccccccccccccc}
\toprule
\multirow{2}{*}{Methods}             & \multicolumn{2}{c}{Hydrant} & \multicolumn{2}{c}{Bench} & \multicolumn{2}{c}{Donut} & \multicolumn{2}{c}{Teddy bear} & \multicolumn{2}{c}{Apple \dag} & \multicolumn{2}{c}{Vase \dag} & \multicolumn{2}{c}{Plant \dag} & \multicolumn{2}{c}{Suitcase \dag} & \multicolumn{2}{c}{Ball \dag} & \multicolumn{2}{c}{Cake \dag}\\ \cmidrule(l){2-3} \cmidrule(l){4-5} \cmidrule(l){6-7} \cmidrule(l){8-9} \cmidrule(l){10-11} \cmidrule(l){12-13} \cmidrule(l){14-15} \cmidrule(l){16-17} \cmidrule(l){18-19} \cmidrule(l){20-21}
                    & \small PSNR   & \small LPIPS   & \small PSNR   & \small LPIPS  & \small PSNR  & \small LPIPS  & \small PSNR   & \small LPIPS & \small PSNR   & \small LPIPS & \small PSNR   & \small LPIPS & \small PSNR   & \small LPIPS & \small PSNR   & \small LPIPS & \small PSNR   & \small LPIPS & \small PSNR   & \small LPIPS    \\ \midrule              
NerFormer \cite{reizenstein2021common}   & 18.2   & 0.30   & 15.9   & 0.43  & 20.2   & 0.34   &  15.8    & 0.44  & 19.5 & 0.33 & 17.7 & 0.34 & 17.8 & 0.45 & 20.0 & 0.39 & 16.8 & 0.35 & 16.9 & 0.44 \\
ViewFormer \cite{kulhanek2022viewformer} & 17.5   & 0.16   & 16.4   & 0.30  & 18.6   & 0.24   &  15.6    & 0.33  & 20.1 & 0.26 & 20.4 & 0.21 & 17.8 & 0.31 & 21.0 & 0.26 & 18.3 & 0.31 & 17.3 & 0.33 \\
SparseFusion \cite{zhou2023sparsefusion} & \underline{22.3}  & \underline{0.16}  & \underline{16.7}   & \underline{0.29} & \textbf{22.8}   & \underline{0.22}   & \underline{20.6}  & \underline{0.24} & \underline{22.8} & \underline{0.20} & \underline{22.8} & \underline{0.18} & \textbf{20.0} & \underline{0.25} & \underline{22.2} & \underline{0.22} & \textbf{22.4} & \underline{0.22} & \underline{20.8} & 0.\underline{28}\\
Ours                                     & \textbf{22.6}  & \textbf{0.15}   &  \textbf{18.4} & \textbf{0.28} & \underline{22.7}  & \textbf{0.22}  &  \textbf{20.7} &  \textbf{0.22} & \textbf{23.0} & \textbf{0.18} & \textbf{22.8} & \textbf{0.16} & \underline{19.0} & \textbf{0.24} & \textbf{22.8} & \textbf{0.21} & \underline{22.2} & \textbf{0.20} & \textbf{20.9} & \textbf{0.28} \\
\bottomrule
\end{tabular}
}
\caption{\textbf{Quantitative comparison in view quality of reconstruction from two views on core-10 categories.} We follow the experiment setting from \cite{zhou2023sparsefusion}, and report PSNR $\uparrow$, and LPIPS $\downarrow$ averaged across the first ten scenes from the testing set. Baselines results marked with `\dag'   
 are reported by \cite{zhou2023sparsefusion}.}
 \label{tab:2views}
\end{table}

\begin{figure}[tb]
  \centering
  \includegraphics[width=1.\linewidth]{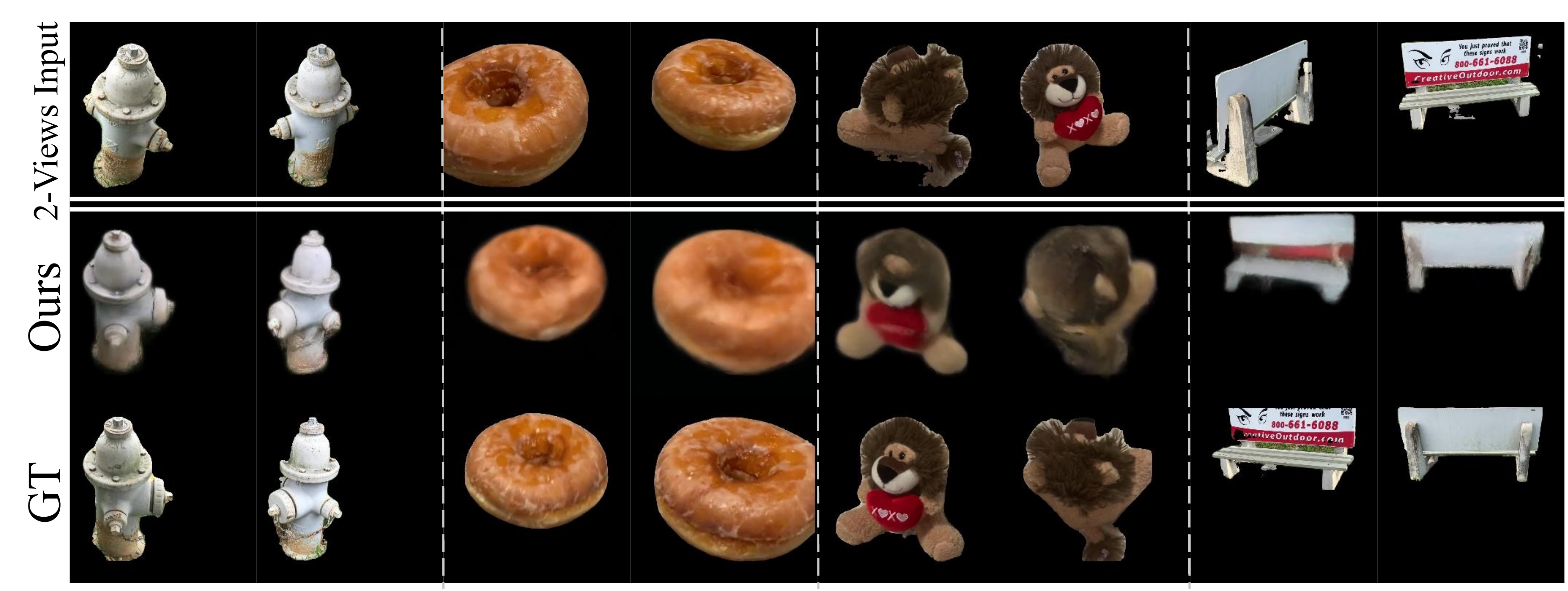}
  \caption{\textbf{View synthesis qualitative results from 2-views reconstruction.} We provide the visual results coherent with \cite{zhou2023sparsefusion} demo setting.}
  \label{fig:results2view2d}
\end{figure}

\begin{figure}[tbh]
  \centering
  \begin{minipage}{0.6\linewidth}
    \centering
    \includegraphics[width=\linewidth]{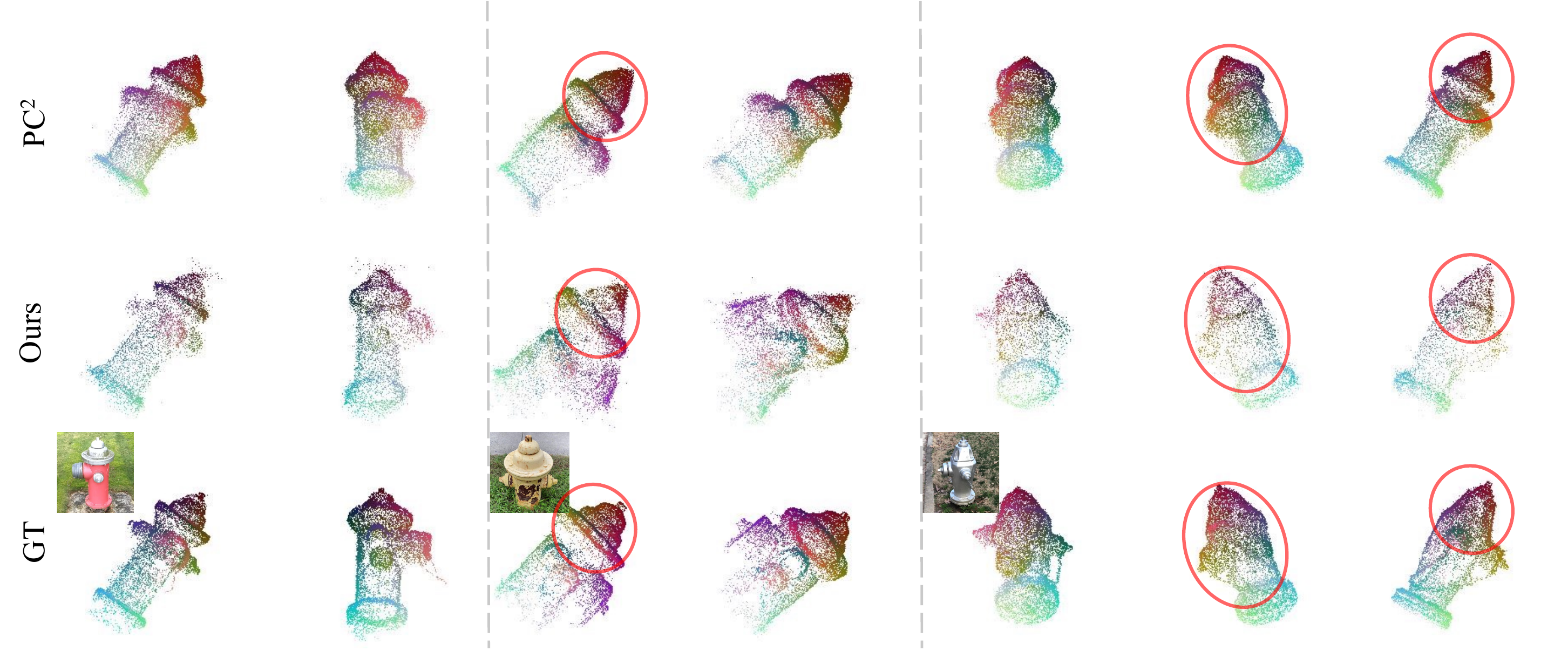}
  \caption{\textbf{Point cloud visualization of single-view reconstruction.} We visualize the object reconstructed from single-view input on hydrant. For our method, the point cloud is extracted from the position feature of GS.}
    \label{fig:comp3dsub}
    \end{minipage}
    \hfill
  \begin{minipage}{0.38\linewidth}
    \centering
\resizebox{1.\linewidth}{!}{ 
\begin{tabular}{lcc}
\toprule
\multirow{2}{*}{Methods}             & \multicolumn{2}{c}{Hydrant}\\ \cmidrule(l){2-3}
                    & F-score   & ChamferDist  \\ 
\midrule
PC\textsuperscript{2} \cite{melas-kyriazi_pc2_2023}  & \underline{0.185}   & 0.073 \\
Ours  & \textbf{0.191}   & \underline{0.068}  \\
Ours w/o iter & 0.180   & \textbf{0.067}  \\
\bottomrule
\end{tabular}
}
\captionof{table}{\textbf{Quantitative comparison in 3D geometry of reconstruction from single view.} We report F-score@0.01 $\uparrow$ and Chamfer Distance $\downarrow$ averaged across 50 scenes in the testing set of hydrant.}
\label{tab:3d}
    \end{minipage}
\end{figure}

\Cref{tab:1view} shows that our approach outperforms other methods on all view synthesis metrics. The performance is better illustrated in the qualitative comparison presented in \Cref{fig:results1view2d}. The baselines either produce blurry views or struggle to maintain the view consistency. Our reconstruction faithfully keeps the content from the given view and generalizes to the whole 3D to generate views with geometrical consistency. 

To further demonstrate the flexibility and efficacy of our view-guided sampling strategy, we additionally report results for reconstruction from 2-views, in \Cref{tab:2views} and qualitatively in \Cref{fig:results2view2d}. By perceiving more views of the object, our approach can gain improved results, very comparable to the most current state-of-the-art, \textit{SparseFusion}. While \textit{SparseFusion} takes much longer time for SDS to extract the 3D scene.

For 3D geometry accuracy, we show quantitative comparison in \Cref{tab:3d}. 
Since the scenes are highly unconstrained, we normalize each object with its ground truth bounding box to evaluate it on the unit scale. While our model not only focuses on 3D geometry accuracy but also takes realistic view synthesis into account, it outperforms the state-of-the-art approach, \textit{i.e.} PC\textsuperscript{2} \cite{melas-kyriazi_pc2_2023}, concentrating on point cloud reconstruction. The qualitative results from \Cref{fig:comp3dsub} further reveal the strength of our approach. PC\textsuperscript{2} tends to reconstruct the object in a category-mean shape, so it struggles when the target deviates a bit more from the dataset distribution center. We argue that it mainly results from how they add conditions to the diffusion model, which we will discuss in \cref{subsec:abla}. Although our results may contain sparse outlier points around the object due to the nature of GS, we achieve a better fit to the 3D shape of various instances.

\subsection{Ablation Studies}
\label{subsec:abla}

\begin{figure}[tbh]
  \centering
  \begin{minipage}{0.45\linewidth}
    \centering
  \includegraphics[width=0.9\linewidth]{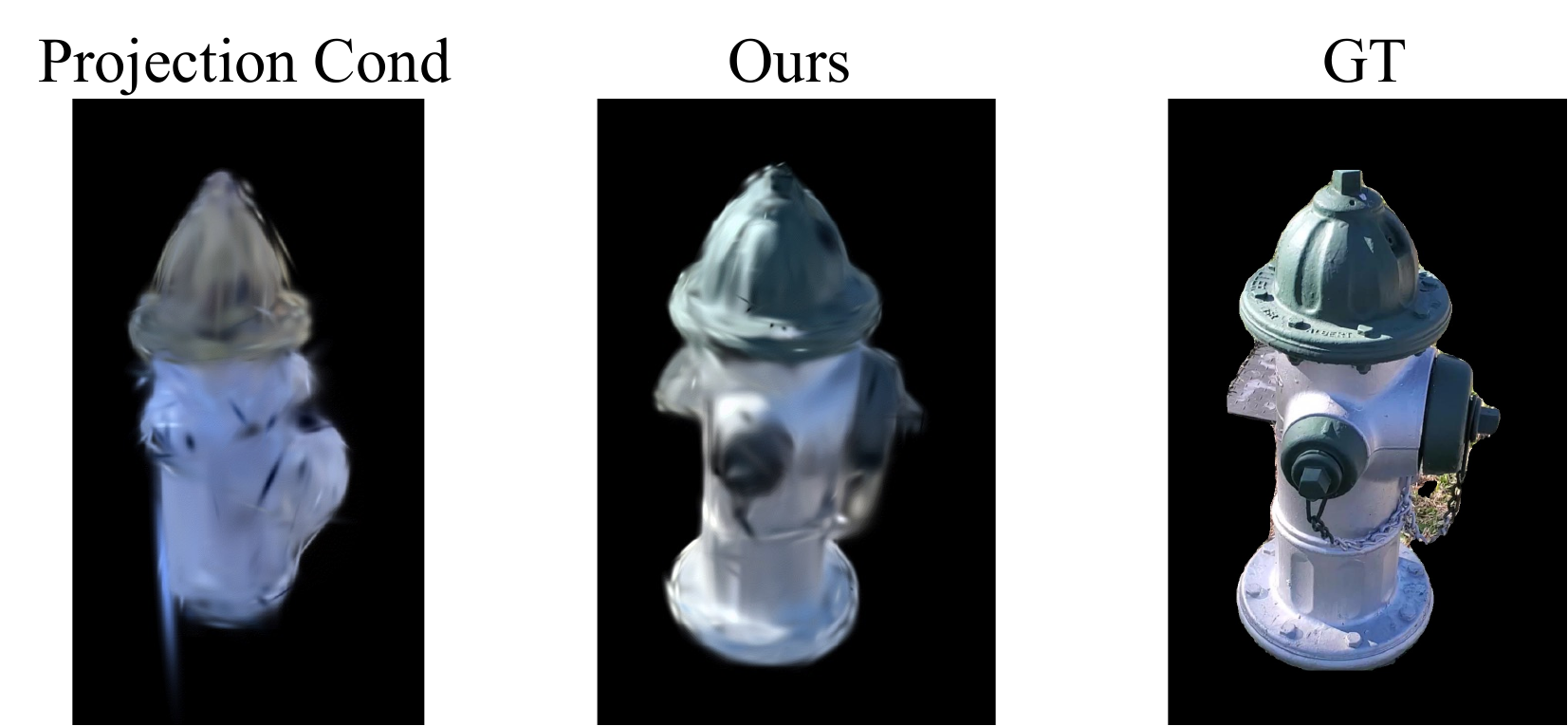}
  \caption{\textbf{Qualitative results on different strategies to add conditions.} For clarity, we present these results without improvement from our iterative polishing and re-using process.}
  \label{fig:ablacondsub}
  \end{minipage}
  \hfill
  \begin{minipage}{0.5\linewidth}
  \centering
    \resizebox{0.8\linewidth}{!}{ 
    \begin{tabular}{lcccc}
    \toprule
    \multirow{2}{*}{Methods}             & \multicolumn{4}{c}{Hydrant}\\ \cmidrule(l){2-5}
                        & PSNR   & SSIM & LPIPS & Time  \\ 
    \midrule
    ProjCond  & 15.10   & 0.519 & 0.407 & \textbf{0.12}s\\
    Ours & \textbf{18.61}   & \textbf{0.783} & \textbf{0.265} & 0.17s\\
    \bottomrule
    \end{tabular}
    }
    \captionof{table}{\textbf{Quantitative comparison of different adding-conditions strategies on single-view reconstruction.} We report PSNR $\uparrow$, SSIM $\uparrow$, LPIPS $\downarrow$, and inference time per denoising step $\downarrow$ averaged across all scenes in the hydrant testing set.}
    \label{tab:ablacond}
  \end{minipage}

\end{figure}

\textbf{Effect of View-Guided Sampling.}
\cref{fig:ablacondsub} compares the different approaches to add conditions to the diffusion model in this task. Both results are rendered from purely GS reconstruction without iterative polishing and re-using process. For the forward projection method, we adopt a fine-grained feature extraction module similar to the point projection condition in PC\textsuperscript{2} \cite{melas-kyriazi_pc2_2023} to train a conditional diffusion model. We utilize classifier-free guidance at inference time to reconstruct the object from the given view. However, empirical results from \cref{fig:comp3dsub} and \cref{fig:ablacondsub} both suggest that this adding-conditions strategy is not as effective as our proposal. Concurrent studies \cite{di2023ccd} also argue that this point projection-based strategy is unstable for relatively special cases.

\begin{figure}[tbh]
\begin{minipage}{0.50\linewidth}
    \centering
    \includegraphics[width=0.95\linewidth]{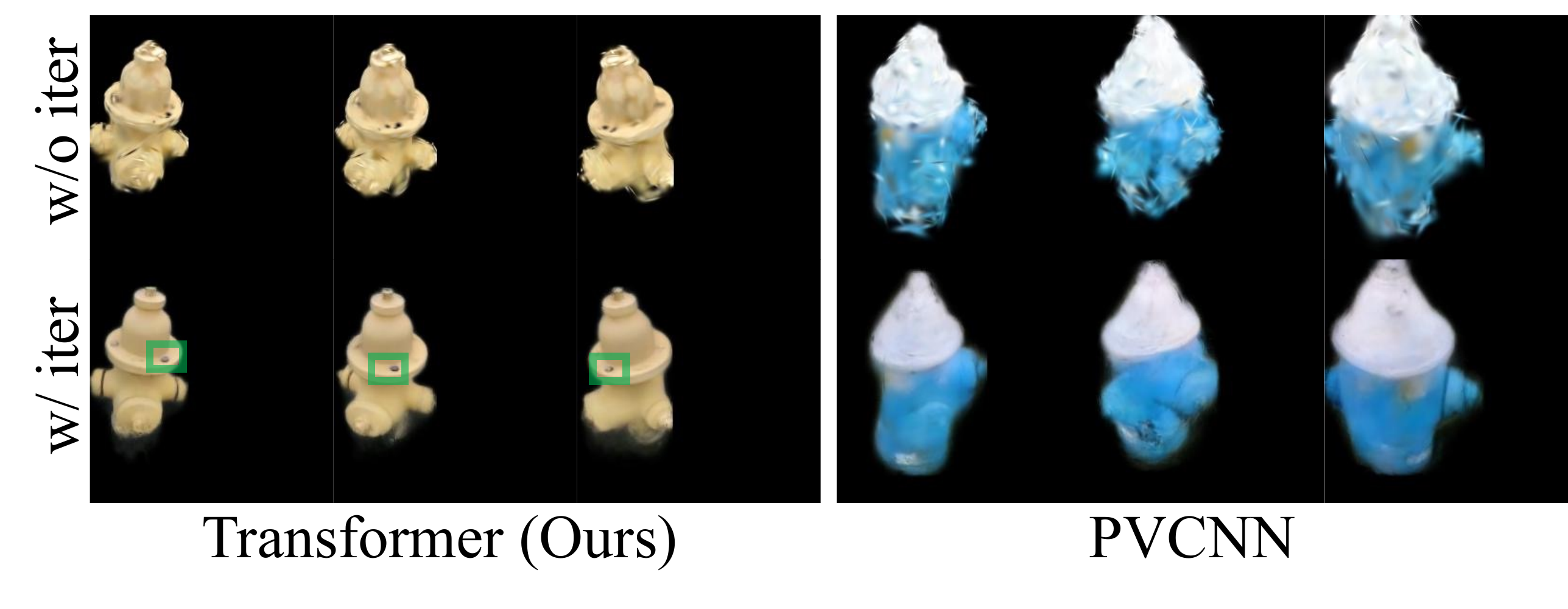} 
      \caption{\textbf{Qualitative results on different backbones for GS denoiser $p_\theta$.} We show unconditional generation results rendered from GS by models built on Transformer and PVCNN~\cite{liu2019point}, w/ and w/o iterative polishing and re-using process.}
      \label{fig:ablabackbone}
\end{minipage}
\hfill
\begin{minipage}{0.47\linewidth}
    \centering
      \includegraphics[width=\linewidth]{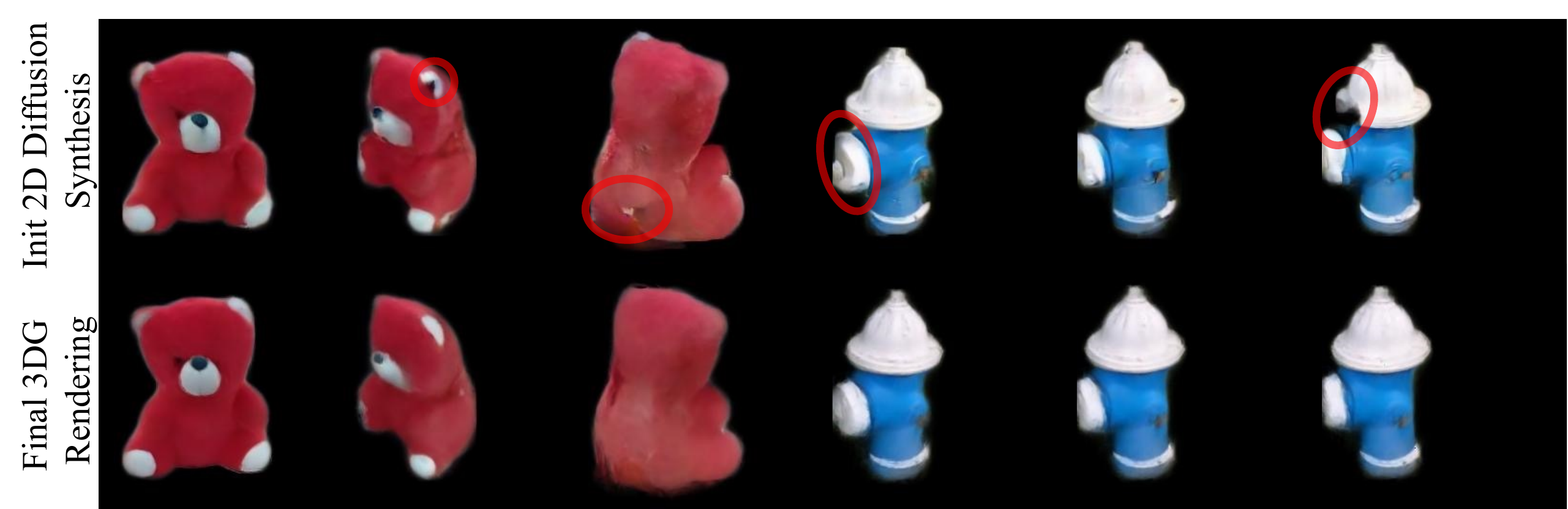}
      \caption{\textbf{Qualitative results of continuous frames from the initial 2D view diffusion and the final reconstructed GS rendering.} The inconsistency and artifacts are emphasized by the \textcolor{red}{red} circle.}
      \label{fig:viewconsisten}
\end{minipage}
\end{figure}

\noindent \textbf{Choice of backbone for GS modeling.}
We compare the transformer we used with PVCNN, a currently popular backbone for unstructured data learning, used by PC\textsuperscript{2} \cite{melas-kyriazi_pc2_2023}, LION \cite{zeng2022lion}. To examine how the model learns the distribution from the dataset, \cref{fig:ablabackbone} shows the unconditional generation results with and without our iterative polishing and re-using  strategy. Both results support that the transformer learns the GS dataset distribution better. This is possible because the transformer organizes edges implicitly while PVCNN uses the explicit point position, which is not suitable for the GS feature since GS attributes exhibit strong correlations. For example, the covariance vectors also contain spatial information in addition to positions.

\noindent \textbf{Effect of Iterative polishing and re-using process.}
Quantitative results from \cref{tab:1view}, suggest the efficacy of our iterative polishing and re-using  strategy in improving the view rendering quality for the reconstructed GS. This is also supported by \cref{fig:ablabackbone} and \cref{fig:viewconsisten}. The \textcolor{green}{green} square highlights the geometry consistency which is inherently the strength of modeling in 3D space. The comparable view quality in these paired results supports that our iterative polishing and re-using process improves reconstructed GS view quality with the assistance of 2D diffusion. On the other hand, modeling in 3D inherently enhances the view consistency for novel view synthesis.

\subsection{Limitations}
The primary limitation of our model is the need for GS ground truth for training. It would limit us from scaling up our model to a common image dataset or working on a generic object reconstruction scenario. We adopted a constrained densification to obtain GS ground truth which has been empirically examined to be relatively efficient while still providing a wide area for exploration.

\begin{figure}[tbh]
  \centering
  \includegraphics[width=1.\linewidth]{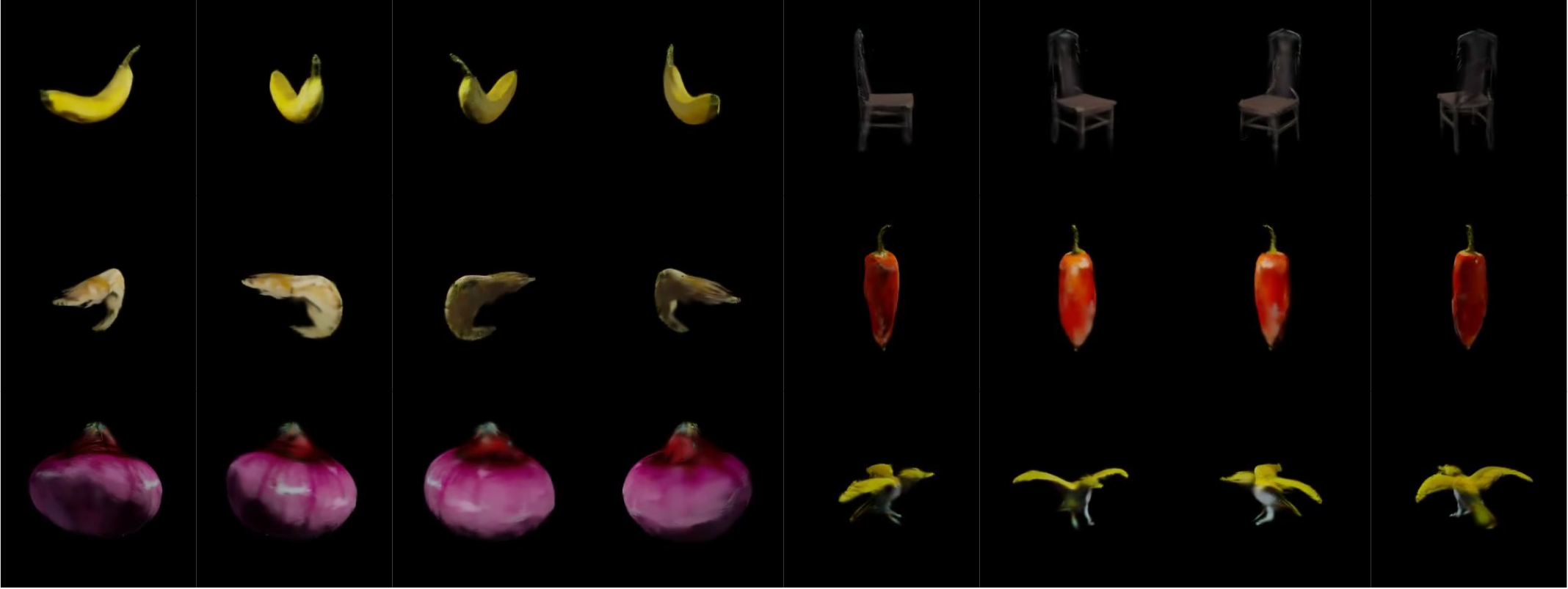}
  \caption{\textbf{Qualitative results of general object reconstruction with a single model trained on OmniObject3D.} }
  \label{fig:omniresults}
\end{figure}

\section{Conclusion}
\label{sec:discussio}
We proposed GSD, a generative real-world object reconstruction approach from a single image using a Diffusion Transformer upon Gaussian Splatting. We  make use of the splatting function for efficient fine-grained 2D feature perception with view-guided sampling. The proposed method has showcased superior performance in category-specific reconstruction tasks. Thanks to the DiT and the fine-grained conditioning mechanism, GSD exhibits the potential to scale up \cref{fig:omniresults}, which could pave the way toward achieving photo-realistic performance in generic object reconstruction tasks.

\newpage
\section*{ACKNOWLEDGEMENTS}
We gratefully acknowledge the support of MindSpore (https://www.mindspore.cn/), CANN (Compute Architecture for Neural Networks) and Ascend AI Processor used for this research. The UAlberta team is partially supported by the NSERC Discovery, the NSERC AICA, the CASBE-Alliance, and the CFI-JELF grants.

%
%
\bibliographystyle{splncs04}
\bibliography{main}

\end{document}